\PassOptionsToPackage{table,xcdraw}{xcolor}
\documentclass[sigconf]{acmart}
\AtBeginDocument{%
  \providecommand\BibTeX{{%
    \normalfont B\kern-0.5em{\scshape i\kern-0.25em b}\kern-0.8em\TeX}}}

\usepackage[linesnumbered,lined,vlined,ruled,commentsnumbered]{algorithm2e}
\usepackage{booktabs}
\usepackage{balance}
\usepackage{multirow}
\usepackage{xcolor}
\usepackage[title,toc,titletoc,page]{appendix}
\usepackage{pifont}
\copyrightyear{2022} 
\acmYear{2022} 
\setcopyright{acmcopyright}\acmConference[MM '22]{Proceedings of the 30th ACM International Conference on Multimedia}{October 10--14, 2022}{Lisboa, Portugal}
\acmBooktitle{Proceedings of the 30th ACM International Conference on Multimedia (MM '22), October 10--14, 2022, Lisboa, Portugal}
\acmPrice{15.00}
\acmDOI{10.1145/3503161.3547853}
\acmISBN{978-1-4503-9203-7/22/10}

\usepackage{soul}
\usepackage[utf8]{inputenc}
\usepackage{graphicx}
\usepackage{amsmath}
\usepackage{amsthm}
\usepackage{multirow}
\usepackage{color}
\newcommand{\lyx}[1]{\textcolor{black}{#1}}
\newcommand{\ljx}[1]{\textcolor{black}{#1}}
\newcommand{\seasons}[1]{\textcolor{black}{#1}}
\begin{document}

%\title{Adaptive Multi Distance Metrics for Few-shot Classification}
\title{Rethinking the Metric in Few-shot Learning: From an Adaptive Multi-Distance Perspective}


% \author{Jinxiang Lai\textsuperscript{1}, Siqian Yang\textsuperscript{1}, Guannan Jiang\textsuperscript{2}, Xi Wang\textsuperscript{2}, Yuxi Li\textsuperscript{1}, Zihui Jia\textsuperscript{1}, Xiaochen Chen\textsuperscript{1}, Jun Liu\textsuperscript{1}, Bin-Bin Gao\textsuperscript{1}, Wei Zhang\textsuperscript{2}, Yuan Xie\textsuperscript{3\dag}, Chengjie Wang\textsuperscript{1\dag}}
% \affiliation{
% \textsuperscript{1}{Tencent Youtu Lab}, 
% \textsuperscript{2}{CATL}, 
% \textsuperscript{3}{School of Computer Science and Technology, East China Normal University
% \country{China}}}
% \email{{jinxianglai, seasonsyang, xibeijia, husonchen, jasoncjwang}@tencent.com}
% \email{{jianggn, WangX30, zhangwei}@catl.com, lyxok1@sjtu.edu.cn, {junsenselee,csgaobb}@gmail.com, yxie@cs.ecnu.edu.cn}
% %\email{}
% \thanks{\dag Corresponding Author}
\settopmatter{authorsperrow=4}
\author{Jinxiang Lai}
\author{Siqian Yang}
\email{jinxiangla@tencent.com}
\email{seasonsyang@tencent.com}
\affiliation{Tencent Youtu Lab \country{China}}

\author{Guannan Jiang}
\author{Xi Wang}
\email{jianggn@catl.com}
\email{wangx30@catl.com}
\affiliation{CATL \country{China}}

\author{Yuxi Li}
\author{Zihui Jia}
\email{lyxok1@sjtu.edu.cn}
\email{xibeijia@tencent.com}
\affiliation{Tencent Youtu Lab \country{China}}

\author{Xiaochen Chen}
\email{husonchen@tencent.com}
\affiliation{Tencent Youtu Lab \country{China}}

\author{Jun Liu}
\author{Bin-Bin Gao}
\email{junsenselee@gmail.com}
\email{csgaobb@gmail.com}
\affiliation{Tencent Youtu Lab \country{China}}

\author{Wei Zhang}
\email{zhangwei@catl.com}
\affiliation{CATL \country{China}}

\author{Yuan Xie\dag}
\email{yxie@cs.ecnu.edu.cn}
\affiliation{School of Computer Science and Technology, East China Normal University \country{China}}

\author{Chengjie Wang\dag}
\email{jasoncjwang@tencent.com}
\affiliation{Tencent Youtu Lab \country{China}}

\thanks{\dag Corresponding Author}

% \author{Jinxiang Lai}
% \author{Siqian Yang}
% \author{Yuxi Li}
% \email{jinxiangla@tencent.com}
% \email{seasonsyang@tencent.com}
% \email{yxie@cs.ecnu.edu.cn}
% \affiliation{Tencent Youtu Lab \country{China}}

% \author{Guannan Jiang}
% \author{Xi Wang}
% \author{Wei Zhang}
% \email{jianggn@catl.com}
% \email{WangX30@catl.com}
% \email{zhangwei@catl.com}
% \affiliation{CATL \country{China}}

% \author{Zihui Jia}
% \author{Xiaochen Chen}
% \author{Bin-Bin Gao}
% \email{xibeijia@tencent.com}
% \email{husonchen@tencent.com}
% \email{csgaobb@gmail.com}
% \affiliation{Tencent Youtu Lab \country{China}}

% \author{Jun Liu}
% \author{Chengjie Wang\dag}
% \email{junsenselee@gmail.com}
% \email{jasoncjwang@tencent.com}
% \affiliation{Tencent Youtu Lab \country{China}}

% \author{Yuan Xie\dag}
% \email{yxie@cs.ecnu.edu.cn}
% \affiliation{School of Computer Science and Technology, East China Normal University \country{China}}

% \thanks{\dag Corresponding Author}
\renewcommand{\shortauthors}{Jinxiang Lai, et al.}

%we show that a good fusion of multiple distance metrics could bring 
\begin{abstract}
Few-shot learning problem focuses on recognizing unseen classes given a few labeled images. In recent effort, more attention is paid to fine-grained feature embedding, ignoring the relationship among different distance metrics. In this paper, for the first time, we investigate the contributions of different distance metrics, and propose an adaptive fusion scheme, bringing significant improvements in few-shot classification. We start from a naive baseline of confidence summation and demonstrate the necessity of exploiting the complementary property of different distance metrics. By finding the competition problem among them, built upon the baseline, we propose an \textit{Adaptive Metrics Module} (AMM) to decouple metrics fusion into metric-prediction fusion and metric-losses fusion. The former  encourages mutual complementary, while the latter alleviates metric competition via multi-task collaborative learning. Based on AMM, we design a few-shot classification framework AMTNet, including the AMM and the \textit{Global Adaptive Loss} (GAL), to jointly optimize the few-shot task and auxiliary self-supervised task, making the embedding features more robust. In the experiment, the proposed AMM achieves $2\%$ higher performance than the naive metrics fusion module, and our AMTNet outperforms the state-of-the-arts on multiple benchmark datasets.
\end{abstract}





\begin{teaserfigure}
\centering
\includegraphics[width=1.0\linewidth]{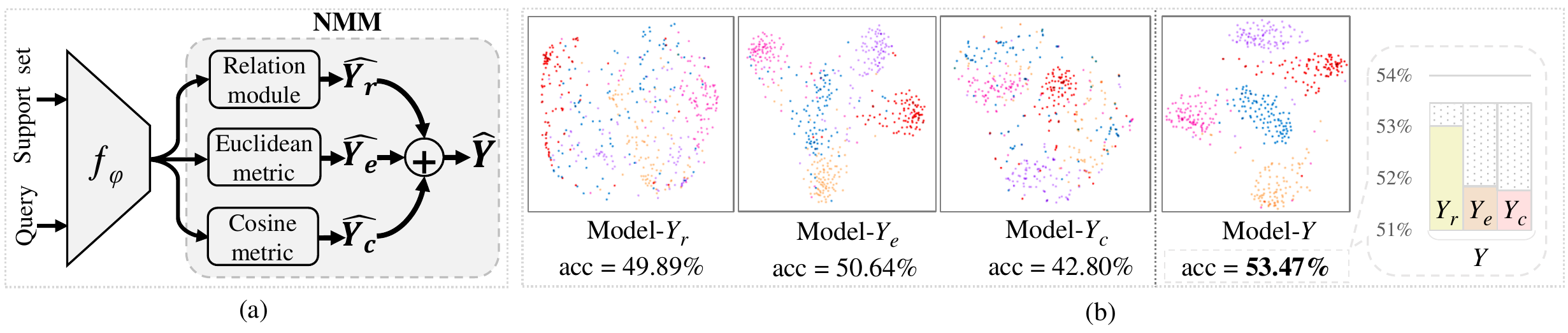}
\vspace{-8mm}
\caption{Effectiveness of different metrics and their combination. {\color{black}(a) The} Model-${Y}$ consists of embedding ${f_\varphi}$ (Conv4) and NMM which integrates three metrics by summing their predictions and more details are in Sec.\ref{subsec:NMM}. {\color{black}(b) Comparison} between different metrics on 5-way 1-shot classification on \emph{mini}ImageNet. The Model-${Y_r}$, Model-${Y_e}$ and Model-${Y_c}$ adopt Relation module, Euclidean and Cosine {\color{black}metrics in separated training way, respectively.}
In right corner, Model-${Y}$ achieves 53.47\%, while the integrated three metrics obtain different accuracy {\color{black}in joint training way (detail accuracy are shown in the last column of Tab.\ref{table:NMM})}.
}
\label{fig:NMM}
\vspace{-0.5mm}
\end{teaserfigure}

\maketitle

\section{Introduction}
\label{sec:Introduction}
\textit{Few-Shot Learning} (FSL) task inspires from the association ability of humans, which tries to learn a transferable classifier given a few samples of each class. 
General FSL methods consist of two parts: feature embedding and distance metrics.
{\color{black}Recent works \cite{tian2020rethinking,snell2017prototypical} have demonstrated that well-trained embedding is helpful to identify samples of the same category, which benefits following distance measurement.
However, the contribution of different distance metrics has not been uniformly studied so far.}

The distance metrics in recent investigations can be divided into two categories: non-parametric fixed distance metric \cite{vinyals2016matching,hou2019cross,snell2017prototypical} (i.e., Euclidean or Cosine metrics), and flexible distance metric \cite{sung2018learning,wu2019parn} with learnable parameters.
To the best of our knowledge, there is no uniform investigation to measure the effect of different distance metrics, therefore, we first conduct empirical analysis over different distance metrics. 
We start with some simple experiments where few-shot learner is trained with three classic distance metrics, including flexible measurement (Relation module \cite{sung2018learning}) and fixed metrics (Cosine and Euclidean distances).
As illustrated in Fig.~\ref{fig:NMM}(b), t-SNE visualizations of the embedding features show that Cosine metric (Model-${Y_c}$) and Euclidean metric (Model-${Y_e}$) {\color{black}learn more discriminative embedding (with clear separation gap) under inter-class constraint provided by \emph{fixed metrics}, while Relation module (Model-${Y_r}$) pay more attention to learn a reasonable embedding structure under intra-instance restriction enforced by \emph{flexible metric}.}

In order to obtain a flexible metric while keeping discriminative embedding, we design a simple metric fusion solution, termed as \textit{Naive Metrics Module} (NMM), which combines three metrics by {\color{black} directly} adding their prediction results. 
In Fig.~\ref{fig:NMM}, {\color{black}the Model-${Y}$ adopts the NMM to achieve a comprehensive decision metric with a more reasonable intra-instance structure (the intra-class distribution tends to uniform).
Meanwhile, the Model-${Y}$ learns a discriminative embedding with the constraints of Cosine and Euclidean metrics}. 
Therefore, the Model-${Y}$ takes advantages of both learnable and fixed metrics to offset the weaknesses when each metric is individually applied.

\begin{figure}[t]
\centering
\includegraphics[width=0.92\linewidth]{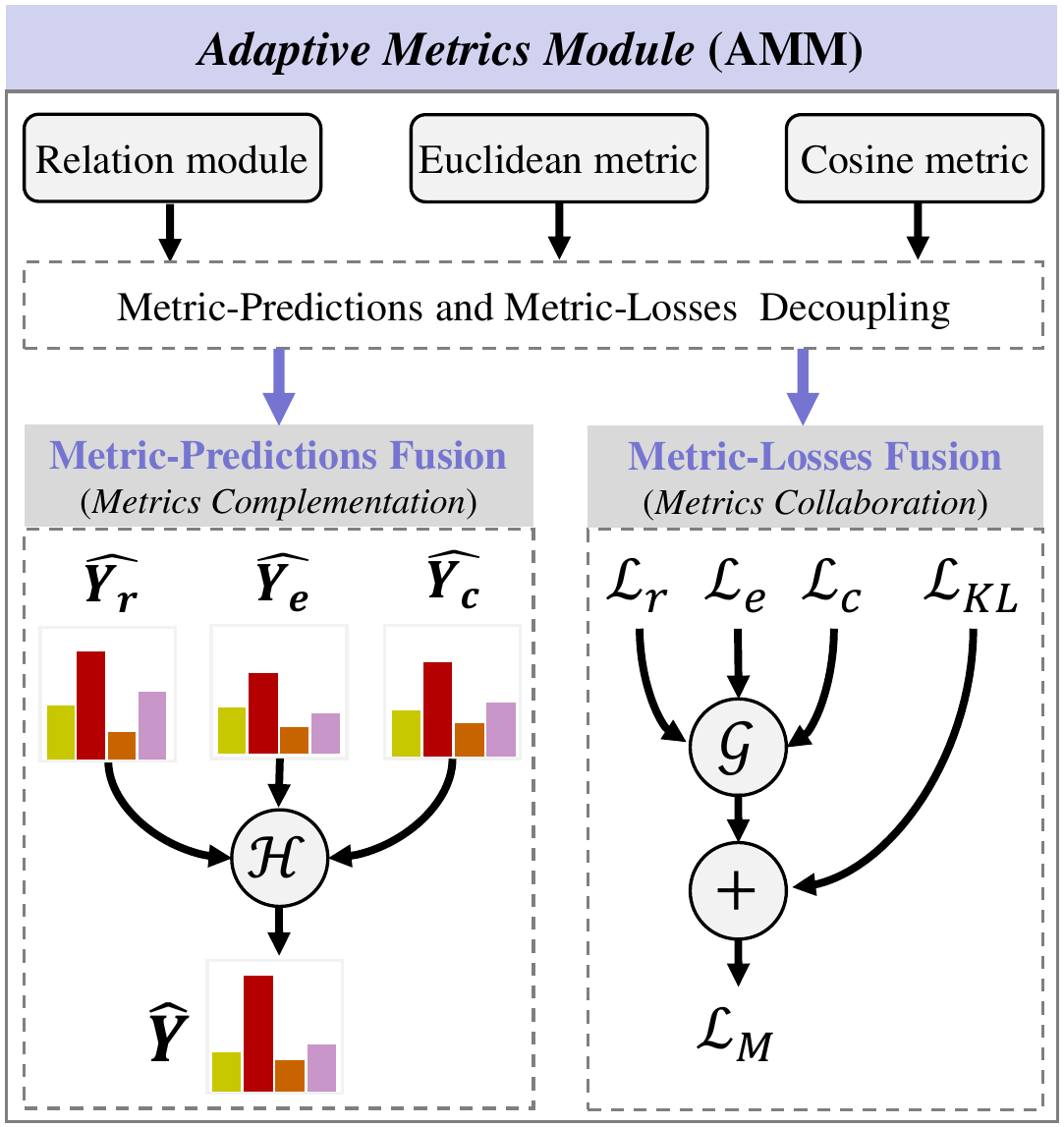}
\vspace{-1mm}
\caption{
%The proposed \textit{Adaptive Metrics Module} (AMM), and more details are described in Sec.\ref{sec:Preliminary} and Sec.\ref{sec:metrics_fusion}. 
\ljx{The proposed \textit{Adaptive Metrics Module} (AMM) decouples metrics fusion into metric-predictions fusion and metric-losses fusion to realize metrics complementation and collaboration respectively, where $\mathcal{H}$ and $\mathcal{G}$ are the functions of predictions fusion and losses fusion respectively.}
}
\label{fig:AMM}
\vspace{-4mm}
\end{figure}

The results in Fig. \ref{fig:NMM} reveals that the fixed learning metric and the flexible learning metric can be complementary.
{\color{black}Furthermore, by comparing results of different combinations between any two metrics (shown in Tab.~\ref{table:NMM}, the detailed analysis is provided in Sec. \ref{subsec:NMM})}, we find there is competition between different distance metrics. 
For example, at the 7th row in Tab.~\ref{table:NMM}, there are differences among the contributions of the cosine metric with different combinations. 
The large contrast is ($51.77\%-43.67\%=8.1\%$).
This means if we apply NMM, the three metrics cannot maximize their effectiveness, as illustrated in the right corner of Fig. \ref{fig:NMM}(b). 

{\color{black}To further explore complementary among different metrics while alleviate the \emph{metrics competition problem}, we proposed the \textit{Adaptive Metrics Module} (AMM) as illustrated in Fig.~\ref{fig:AMM}.
The AMM inserts adaptive layers to automatically learn the weights of different metrics via considering them as multiple metric-learning tasks, instead of establishing a uniform standard for measurement.}
Specifically, the AMM decouples metrics fusion into metric-predictions fusion and metric-losses fusion:
(i) The metric-predictions fusion utilizes an adaptive layer to re-weight the contributions of different metrics (i.e. \seasons{$\{\seasons{\hat{Y}}_r,\seasons{\hat{Y}}_e,\seasons{\hat{Y}}_c\}$}, which are the metric-predictions of \{\textit{Relation module, Euclidean metric, Cosine metric}\} respectively);
(ii) The metric-losses fusion guides the model to learn a generalized-well embedding via multi-task collaborative learning paradigm (i.e. $\mathcal{G}\left(\mathcal{L}_r,\mathcal{L}_e,\mathcal{L}_c\right)$). Besides, a KL regularization term $\mathcal{L}_{KL}$ is added to increase the consistency between predictions of each metric.

{\color{black}Based on the proposed AMM and inspired by \cite{tian2020rethinking,rizve2021exploring}, we establish a framework, named \textit{Adaptive Metrics and Tasks Network} (AMTNet), as illustrated in Fig.~\ref{fig:AMTNet}, to integrate auxiliary self-supervised tasks for FSL.}
To maximize the performance, a \textit{Global Adaptive Loss} (GAL) is designed in the framework, which refs to Pareto Optimal multi-task collaborative learning \cite{zhengyu2021pareto}, to merge the embedding from auxiliary tasks and main few-shot classification task.
In addition, AMTNet utilizes the GAL to optimize the whole model in an end-to-end manner. {\color{black}To summarize, our main contributions are: 

\vspace{-0.3mm}
$\bullet$ For the first time, we rethink the role of different types of metric in FSL, and propose to boost the performance from an adaptive multi-distance perspective.

\vspace{-0.3mm}
$\bullet$ By finding the complementary and competition among different metrics, a novel \textit{Adaptive Metrics Module} (AMM) is proposed to integrate the flexible and fixed distance metrics to achieve mutual complementarity. Meanwhile, the collaboration between metrics can be ensured by considering them as multi-task learning.

\vspace{-0.3mm}
$\bullet$ An effective few-shot classification framework AMTNet is designed based on AMM, which leverages a \textit{Global Adaptive Loss} (GAL) to combine few-shot task with auxiliary self-supervised tasks, realizing an end-to-end training.

$\bullet$ AMTNet achieves the state-of-the-art results on multiple benchmark datasets, and the effectiveness of the proposed AMM and GAL is also demonstrated in the experiments.} 

\vspace{-2mm}
\section{Related Work}
\noindent\textbf{Few-Shot Learning:}
FSL algorithms pre-train a base classifier with abundant samples, then learn to recognize novel classes with a few labeled samples.
There are four representative directions of inductive FSL algorithms as briefly introduced in following.

\emph{Optimization-based methods} \cite{ravi2016optimization,marcin2018learn,finn2017model} are able to perform rapid adaption with a few training samples for new classes.
\emph{Parameter-generating methods} \cite{munkhdalai2018rapid,gidaris2019generating} focus on learning a parameter generating network. 
\emph{Embedding-based methods} \cite{tian2020rethinking,rizve2021exploring,zhengyu2021pareto,zhiqiang2021partial,liu2021learning} aim to learn a generalize-well embedding with supervised or self-supervised learning tasks at first, then freeze this embedding and further train a linear classifier or design a metric classifier on novel classes.

\emph{Metric-learning based methods} classify a new input image by computing the similarity compared with labeled instances \cite{gregory2015siamese}.
To learn comparison models, metric-learning based methods make predictions conditioned on distance metrics to few labeled samples during the training stage. 
There are four popular distance metrics: Cosine similarity \cite{vinyals2016matching,hou2019cross,wang2020cooperative}, Euclidean distance \cite{snell2017prototypical}, CNN-based relation module \cite{sung2018learning}, and Earth Mover’s Distance (EMD) \cite{zhang2020deepemd}. 
These methods design carefully on the embedding network to match their corresponding distance metrics.
In this paper, we first investigate the relationships of different distance metrics, and prove that an adaptive fusion brings significant improvements in few-shot classification.

\noindent\textbf{Auxiliary Task in FSL:}
Some recent works gain a performance improvement by training few-shot models with supervised and self-supervised auxiliary tasks.
The supervised task for FSL simply performs global classification on the base dataset as in \cite{hou2019cross}. 
Recently, the effectiveness of self-supervised learning for FSL has been demonstrated in \cite{carlos2020self,doersch2020crosstransformers,liu2021learning,gidaris2019boosting,su2020does,rizve2021exploring}.
In \cite{carlos2020self,liu2021learning}, contrastive learning is employed to improve the generalization ability of embedding features.
In \cite{gidaris2019boosting,su2020does}, an additional rotation prediction task was adopted as auxiliary task to learn more robust features.

In contrast to the existing FSL approaches applied supervised or self-supervised auxiliary tasks, we propose to jointly optimize the main few-shot task and auxiliary tasks in an end-to-end manner. Specifically, our approach adopts AMM based metric classification for main few-shot task, global classification \cite{hou2019cross} for supervised task, and the widely-used and powerful rotation classification for self-supervised task.

\vspace{-3mm}
\section{Preliminary}
\label{sec:Preliminary}
\subsection{Problem Definition}
{\color{black}A few-shot classification usually adopts the ${N}$-way ${K}$-shot episode training strategy, which learns a classifier for $N$ unseen classes with $K$ labeled samples. 
It involves} two mutually disjoint datasets ${X^{base}}$ and ${X^{novel}}$, where the sufficient labeled base set ${X^{base}}$ contains ${C^{base}}$ categories, the few labeled novel set ${X^{novel}}$ has ${C^{novel}}$ categories, and ${C^{base} \cap C^{novel} = \emptyset}$.
In few-shot testing, a set of episodes $\mathcal{T}=\left\{\mathcal{T}_{i}\right\}_{i=1}^{n_e}$ are sampled from ${X^{novel}}$, and the average accuracy over $\mathcal{T}$ are utilized to evaluate the performance of FSL algorithm.
An episode $\mathcal{T}_{i}$ is considered as a ${N}$-way ${K}$-shot task, which contains $N$ classes with $K$ samples per class as the support set $\mathcal{S}=\left\{\left(x^s_i, y^s_i\right)\right\}_{i=1}^{n_s}$ ($n_s=N\times K$), and a fraction of the rest samples as the query set $\mathcal{Q}=\left\{\left(x^q_i, y^q_i\right)\right\}_{i=1}^{n_q}$ ($n_q=N\times T$). 
The support subset of {\color{black}the $k$-th} class is denoted as $\mathcal{S}^k$.
Following \cite{vinyals2016matching,sung2018learning,hou2019cross,xu2021learning}, we adopt the episodic training strategy to mimic the few-shot testing setting.
In particular, the episodic training iteratively samples the same sized episode from base set ${X^{base}}$ to train a meta-learner (i.e., a few-shot classification model). 
After training on ${X^{base}}$, given $N$ unseen classes with $K$ labeled samples, the meta-learner aims to classifier $n_q=N\times T$ unlabeled samples into $N$ categories correctly.

\vspace{-1mm}
\subsection{Metric Classifier}
Metric classifier categorizes the query images into $N$ novel classes based on similarity measurement. 
As shown in Fig.~\ref{fig:NMM}(a), firstly the embedding ${f_\varphi}$ transfers the support set $\mathcal{S}$ and a query sample ${x}^q$ into prototype feature map $P^k=\frac{1}{|\mathcal{S}^k|} \sum_{x^s_i\in \mathcal{S}^k} {f_\varphi}(x^s_i)$ and a query feature map ${Q}={f_\varphi}({x}^q)\in \mathbb{R}^{c\times h\times w}$, respectively. Then, each pair ($P^k$, ${Q}$) is fed into metric classifier to calculate the similarity for classification.
We use $\{d_j\}_{j=r,e,c}$ to denote the corresponding Relation module \cite{sung2018learning}, Euclidean and Cosine metrics, and define $\{\seasons{\hat{y}}_j, \seasons{\hat{Y}}_j, \mathcal{L}_j\}_{j=r,e,c}$ to represent the corresponding metric prediction probability, prediction distribution and loss, respectively. Formally, for metric ${d_j}$, the probability that ${Q}$ belongs to the $k$-th class is:
\begin{equation}
\seasons{\hat{y}}^k_j(Q)=\seasons{\hat{y}}_j(y=k|{Q})=\frac{\exp{\left(-d_j\left({Q}, {P}^k\right)\right)}}
{\sum_{i=1}^{N} \exp{\left(-d_j\left({Q}, {P}^i\right)\right)}}.
\label{equ:pred}
\end{equation}
The individual metric prediction distribution is expressed as:
\begin{equation}
\seasons{\hat{Y}}_j = [\seasons{\hat{y}}^1_j,\dots,\seasons{\hat{y}}^k_j,\dots,\seasons{\hat{y}}^N_j].
\label{equ:Y_vec}
\end{equation}
According to the true $N$-way few-shot class label ${y}^q$, the individual metric classification loss for ${d_j}$ is then defined as:
\begin{equation}
\mathcal{L}_j = CE(\seasons{\hat{y}}_j, {y}^q) = -\sum_{i=1}^{n_q} \log \seasons{\hat{y}}_j(y={y}^q_i|{Q}_{i}),
\label{equ:Lj}
\end{equation}
where ${CE}$ is the cross-entropy loss function.

\vspace{-1mm}
{\color{black}\subsection{Metrics Fusion Methodology}} \label{def_metric_fusion}
In this paper, we first propose metric fusion to construct a compound distance metric for FSL via merging the contributions of different distance metrics. 
{\color{black}The challenge is that, there is no uniform criterion for different metrics. 
To deal with the problem, we decouple metrics fusion step into two aspects (as shown in Fig.\ref{fig:AMM}):} predictions fusion and losses fusion. {\color{black}The former} obtains a comprehensive metric-based predictor by inserting an adaptive re-weighting layer, and {\color{black}the latter} guides the model to learn generalized embedding via multi-task collaborative learning paradigm. 

Formally, the predictions fusion is defined as:
\begin{equation}
\seasons{\hat{Y}} = \mathcal{H}(\seasons{\hat{Y}}_r,\seasons{\hat{Y}}_e,\seasons{\hat{Y}}_c),
\label{equ:pf}
\end{equation}
where $\seasons{\hat{Y}}$ is the overall metric prediction distribution, and $\mathcal{H}$ is the predictions fusion function which aims to tackle the problem of metric criterion discordance. In $\seasons{\hat{Y}}$, {\color{black}the corresponding predicted probability} for the sample $x_q$ can be represented as $\seasons{\hat{y}}$, and its classification loss $\mathcal{L}_y$ is calculated by Eq.~\ref{equ:Lj} as:
\begin{equation}
\mathcal{L}_y = CE(\seasons{\hat{y}}, {y}^q). 
\label{equ:ly}
\end{equation}

Then, the generic losses fusion is defined as:
\begin{equation}
\mathcal{L}_M = \mathcal{F}\left(\mathcal{G}\left(\mathcal{L}_r,\mathcal{L}_e,\mathcal{L}_c\right), \mathcal{L}_y\right),
\label{equ:lf}
\end{equation}
where, $\mathcal{L}_M$ is the overall metric classification loss, $\mathcal{F}$ and $\mathcal{G}$ are losses fusion functions. {\color{black}By treating the losses fusion as a multi-task collaborative learning paradigm, our method avoids the problem of metric criterion discordance, which is helpful for embedding generalization. }

\begin{figure*}[!t]
\centering
\includegraphics[width=0.9\linewidth]{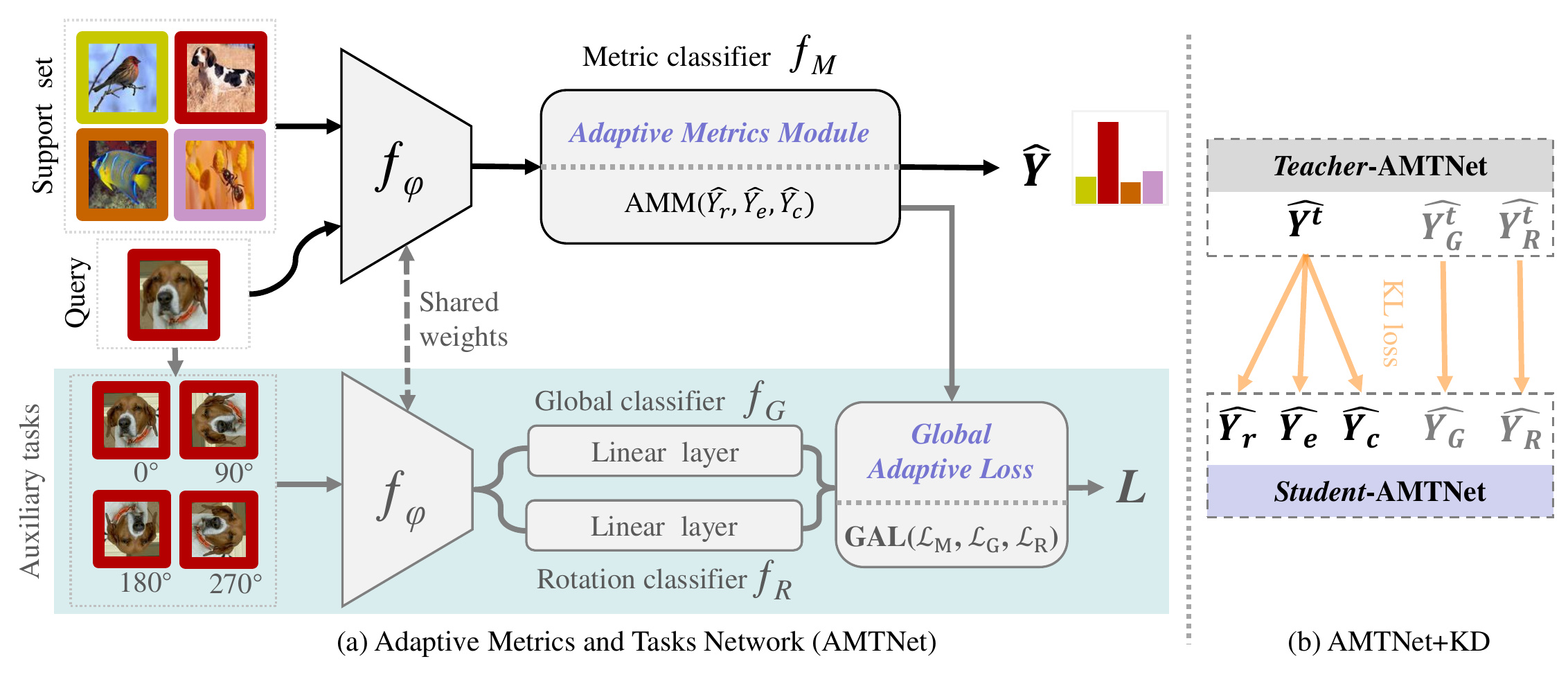}
\vspace{-4.5mm}
\caption{{\color{black}
(a) The framework of the proposed AMTNet.
(b) AMTNet+KD applies knowledge distillation.}}
\label{fig:AMTNet}
\vspace{-3mm}
\end{figure*}

\vspace{-1mm}
\section{Metrics Fusion Model}
\label{sec:metrics_fusion}
In this section, we first introduce a Naive Metrics Module (NMM) to demonstrate the advantages of metrics fusion.
Then, a well-designed metrics fusion approach, named Adaptive Metrics Module (AMM) is proposed based on NMM to further improve the performance.

\vspace{-1mm}
\subsection{Naive Metrics Module}
\label{subsec:NMM}
\textit{Naive Metrics Module} is a naive metric fusion method, which consists of learnable and fixed distance metrics including Relation module, Euclidean and Cosine metrics (illustrated in Fig.~\ref{fig:NMM}(a)). Formally, NMM is expressed as:
\begin{equation}
\seasons{\hat{Y}} = \seasons{\hat{Y}}_r + \seasons{\hat{Y}}_e + \seasons{\hat{Y}}_c,
\label{equ:pf_nmm}
\end{equation}
\begin{equation}
\mathcal{L}_M = \mathcal{L}_r + \mathcal{L}_e + \mathcal{L}_c.
\label{equ:lf_nmm}
\end{equation}
Empirically, the advantage of this simple metric fusion scheme can be reflected in two aspects. First, the metric diversity is enhanced, thus benefit the simiarity measurement. Second, the fusion process naturally exploit the complementarity between learnable and fixed distance metrics.

To evaluate the effectiveness of NMM, we compare the t-SNE of Model-${Y}$, Model-${Y_r}$, Model-${Y_e}$ and Model-${Y_c}$, as illustrated in Fig.~\ref{fig:NMM}. 
We observe that: {\color{black}(i) The embedding feature of the Model-${Y}$ is still discriminative even in the fixed metrics combination.} (ii) The Model-${Y}$ achieves a substantial accuracy improvement compared with the independent counterpart, indicating that integrating the learnable and the fixed metrics achieves mutual complementarity.

To demonstrate the observations of t-SNE visualizations, we conducted more experiments about the different combinations of multiple metrics, and the results shown in Tab.~\ref{table:NMM} indicate that:
(i) With the increase of the metric diversity, the Merge Acc keeps going up.
(ii) The Merge Acc is better than the corresponding Individual Acc, which indicates that integrating multiple metrics can achieve a more comprehensive decision maker. 
For example, in column 8 of Tab.~\ref{table:NMM}, the Merge Acc of Model-${Y}$ achieves 53.47\% which is higher than the corresponding three Individual Acc (53.02\%, 51.85\% and 51.77\% respectively).
(iii) The Individual Acc of the integrating model {\color{black}is superior to the corresponding independent model, and the advantages are more obvious along with the increase of the metric diversity.}
Specifically, comparing column 8 with columns 2, 3 and 4 of Tab.~\ref{table:NMM}, all the Individual Acc of Model-${Y}$ are superior than the corresponding independent Model-${Y_c}$, Model-${Y_e}$ and Model-${Y_r}$. 
Their Individual Acc increase from 42.80\%, 50.64\% and 49.89\% to 51.77\%, 51.85\% and 53.02\%, which indicates that the Model-${Y}$ obtains a more discriminative embedding than these corresponding independent models. 

\renewcommand{\tabcolsep}{2pt}
\begin{table}[ht]
\caption{Comparison between different combinations of multiple metrics on 5-way 1-shot classification on \emph{mini}ImageNet.
The embedding backbone ${f_\varphi}$ is Conv4.
For each column, the model is optimized by $\mathcal{L}_M$ and inferences with different metrics, i.e. each column has the same learned embedding ${f_\varphi}$ for different metrics.
In Individual Acc, each row has the same metric while with different learned embedding.}
\label{table:NMM}
\vspace{-3mm}
\resizebox{8.5cm}{!}
{
\begin{tabular}{cllllllll}
%\hline
\multicolumn{1}{c}{\quad \tiny Column Index} &\multicolumn{1}{c}{\tiny 1} &\multicolumn{1}{c}{\tiny 2} &\multicolumn{1}{c}{\tiny 3} &\multicolumn{1}{c}{\tiny 4} &\multicolumn{1}{c}{\tiny 5} &\multicolumn{1}{c}{\tiny 6} &\multicolumn{1}{c}{\tiny 7} &\multicolumn{1}{c}{\tiny 8} \\
\hline%\noalign{\smallskip}
\multicolumn{2}{c|}{Metric num} & \multicolumn{1}{c}{\small 1} & \multicolumn{1}{c}{\small 1} & \multicolumn{1}{c}{\small 1} & \multicolumn{1}{c}{\small 2} & \multicolumn{1}{c}{\small 2} & \multicolumn{1}{c}{\small 2} & \multicolumn{1}{c}{\textbf{\small 3}}\\
\hline
%\noalign{\smallskip}
\multicolumn{1}{c|}{\multirow{3}*{$\mathcal{L}_M$}}&\multicolumn{1}{c|}{$\mathcal{L}_r$}& \multicolumn{1}{c}{\small \checkmark}&\multicolumn{1}{c}{-}&\multicolumn{1}{c}{-}& \multicolumn{1}{c}{-} & \multicolumn{1}{c}{\small \checkmark} &\multicolumn{1}{c}{\small \checkmark}& \multicolumn{1}{c}{\small \checkmark}\\
\multicolumn{1}{c|}{}&\multicolumn{1}{c|}{$\mathcal{L}_e$} & \multicolumn{1}{c}{-}&\multicolumn{1}{c}{\small \checkmark}&\multicolumn{1}{c}{-}& \multicolumn{1}{c}{\small \checkmark} & \multicolumn{1}{c}{-} &\multicolumn{1}{c}{\small \checkmark}& \multicolumn{1}{c}{\small \checkmark}\\
\multicolumn{1}{c|}{}&\multicolumn{1}{c|}{$\mathcal{L}_c$} & \multicolumn{1}{c}{-}&\multicolumn{1}{c}{-}&\multicolumn{1}{c}{\small \checkmark}& \multicolumn{1}{c}{\small \checkmark} & \multicolumn{1}{c}{\small \checkmark} &\multicolumn{1}{c}{-}& \multicolumn{1}{c}{\small \checkmark}\\
\hline
%\noalign{\smallskip}
\multicolumn{1}{c|}{\multirow{2}*{Individual}}&\multicolumn{1}{c|}{$\seasons{\hat{Y}}_r$} &\multicolumn{1}{c}{\small 49.89}&\multicolumn{1}{c}{-}& \multicolumn{1}{c}{-} &\multicolumn{1}{c}{-}& \multicolumn{1}{c}{\small 51.07} & \multicolumn{1}{c}{\small 50.88} & \multicolumn{1}{c}{\textbf{\small 53.02}}\\
\multicolumn{1}{c|}{\multirow{2}*{Acc}}&\multicolumn{1}{c|}{$\seasons{\hat{Y}}_e$} &\multicolumn{1}{c}{-}& \multicolumn{1}{c}{\small 50.64} &\multicolumn{1}{c}{-}& \multicolumn{1}{c}{\small 51.06} &\multicolumn{1}{c}{-}& \multicolumn{1}{c}{\small 49.63} & \multicolumn{1}{c}{\textbf{\small 51.85}}\\
\multicolumn{1}{c|}{}&\multicolumn{1}{c|}{$\seasons{\hat{Y}}_c$} & \multicolumn{1}{c}{-}&\multicolumn{1}{c}{-}&\multicolumn{1}{c}{\small 42.80}& \multicolumn{1}{c}{\small 51.01} & \multicolumn{1}{c}{\small 43.67} &\multicolumn{1}{c}{-}& \multicolumn{1}{c}{\textbf{\small 51.77}}\\
\hline
\multicolumn{1}{c|}{Merge Acc}&\multicolumn{1}{c|}{$\seasons{\hat{Y}}$} & \multicolumn{1}{c}{\small 49.89} & \multicolumn{1}{c}{\small 50.64} & \multicolumn{1}{c}{\small 42.80} & \multicolumn{1}{c}{\small 51.69} & \multicolumn{1}{c}{\small 51.48} & \multicolumn{1}{c}{\small 52.02} & \multicolumn{1}{c}{\textbf{\small 53.47}}\\
\hline
\end{tabular}
}
\vspace{-4mm}
\end{table}

\ljx{Though NMM is demonstrated to be effective for FSL, the results in Tab.~\ref{table:NMM} \lyx{still} reveals the \lyx{problem of competition} \seasons{among} different distance metrics, i.e. the contributions of metrics are different \lyx{and one metric may bring negative effect to other contributors.} \seasons{In the training stage, different distance metrics have different quantities of their metric losses, which leads to inconsistent gradients in the process of back propagation.
Consequently, competition occurs while training with different distance metrics.}
With the help of the descriptions in section \ref{def_metric_fusion}, the problem can be subdivided into metric criterion discordance in predictions fusion (Eq.~\ref{equ:pf_nmm}), and tasks competition in losses fusion (Eq.~\ref{equ:lf_nmm}).}

\vspace{-1mm}
\subsection{Adaptive Metrics Module}
\ljx{To handle the metrics competition problem,
we propose an \textit{Adaptive Metrics Module} (AMM) on basis of NMM.
The AMM decouples metrics fusion into metric-predictions fusion and metric-losses fusion to realize metrics complementation and collaboration respectively.
}

\noindent\textbf{Metric-Predictions Fusion:}
To deal with the problem of metric criterion discordance in \textit{predictions fusion}, an adaptive layer is inserted in AMM to automatically learn the weights of different metrics:
\begin{equation}
\begin{aligned}
\seasons{\hat{Y}} = \sum_{j=r,e,c}{(1 + u_{j})}{\seasons{\hat{Y}}_j},
\end{aligned}
\label{equ:AMM_Y}
\end{equation}
where ${u_{j}}$ is a learnable variable which is used as a scaling factor reflecting the contribution of $\seasons{\hat{Y}}_j$, and the residual weighted (i.e., $1 + u_{j}$) strategy is applied to ensure learning stability. Then the metric classification loss $\mathcal{L}_y$ for ${\seasons{\hat{Y}}}$ can be calculated by Eq.~\ref{equ:ly}.

\noindent\textbf{Metric-Losses Fusion:}
To tackle {\color{black}the competition problem among different losses}, AMM learns the metric loss weights for the corresponding metric losses $\{{\mathcal{L}_r},{\mathcal{L}_e},{\mathcal{L}_c}\}$. 
Inspired by \cite{alex2018multi}, we use task-dependent uncertainty as a basis to modulate multi-task losses:
\begin{equation}
\begin{aligned}
\mathcal{G}\left(\mathcal{L}_r,\mathcal{L}_e,\mathcal{L}_c\right) &\!=\! -\sum_{i=1}^{n_q} \log( \seasons{\hat{y}}_r \cdot \seasons{\hat{y}}_e \cdot \seasons{\hat{y}}_c ) \approx \sum_{j=r,e,c}({{\frac{1}{{\theta_j^2}}}{\mathcal{L}_j}+{log{\theta_j^2}}}),
\end{aligned}
\label{equ:AMM_Lm}
\end{equation}
where ${\theta_j}$ is a learnable variable for metric $d_j$, and the detailed formula derivation is presented in the APPENDIX.
According to Eq.~\ref{equ:AMM_Lm}, large scale value ${\theta_j}$ will decrease the contribution of ${\mathcal{L}_j}$, whereas small scale ${\theta_j}$ will increase its contribution. The loss ${\mathcal{L}_M}$ is penalized when setting ${\theta_j}$ too small, therefore, {\color{black}it can prevent trivial solutions where some loss terms are degraded to zero.}

Moreover, a KL regularization term $\mathcal{L}_{KL}$ is added to increase consistencies between different metric distributions.
Formally, the overall metric-losses fusion of AMM is expressed as:
\begin{equation}
\begin{aligned}
\mathcal{L}_M &= \mathcal{G}\left(\mathcal{L}_r,\mathcal{L}_e,\mathcal{L}_c\right) + \alpha \mathcal{L}_{KL} \\
&=\sum_{j=r,e,c}({{\frac{1}{{\theta_j^2}}}{\mathcal{L}_j}+{log{\theta_j^2}}}) + \alpha \sum_{j=r,e,c}KL\left(\seasons{\hat{Y}}_j,\|{\seasons{\hat{Y}}}\|\right),
\end{aligned}
\label{equ:AMM_LM}
\end{equation}
{\color{black}where $\alpha$ is a hyper-parameter, $KL(\cdot,\cdot)$ is Kullback–Leibler divergence function.}
\seasons{In detail,} the KL regularization $\mathcal{L}_{KL}$ considers \seasons{the fused prediction} $\seasons{\hat{Y}}$ as the teacher-metric and \seasons{the individual metric predictions} $\seasons{\hat{Y}}_j$ as the student-metric \seasons{in Eq. \ref{equ:AMM_LM}}, which increases consistencies between different metric distributions to alleviate the metric criterion discordance problem.

\vspace{-4mm}
\begin{algorithm}[ht]
\caption{AMTNet model training}
\label{alg:AMTNet}
\SetAlgoLined
\SetKwInput{KwData}{Input}
\SetKwInput{KwModel}{Model}
\SetKwInput{KwResult}{Output}
 \KwData{${X^{base}}$; training epochs $E$}
 \KwModel{Backbone ${f_\varphi}$; Metric classifier (i.e. AMM) ${f_M}$; Global classifier ${f_G}$; Rotation classifier ${f_R}$}
 \KwResult{${f_\varphi}$; ${f_M}$}
 \Begin{
 Randomly initialize \{${f_\varphi}$,${f_M}$,${f_G}$,${f_R}$\}\;
 \For{i \textbf{from} 1 \textbf{to} $E$}{
 Sample training data $(\mathcal{S},\mathcal{Q}) \in {X^{base}}$\;
 Compute $\mathcal{L}$ by Eq.~\ref{equ:Loss} involved $\{\mathcal{L}_M,\mathcal{L}_G,\mathcal{L}_R\}$\;
 Optimize \{${f_\varphi}$,${f_M}$,${f_G}$,${f_R}$\} with SGD\;
 Freeze all params except $\{u_r,u_e,u_c\}$ in Eq.~\ref{equ:AMM_Y}\;
 Compute loss $\mathcal{L}_y$ by Eq.~\ref{equ:ly}\;
 Optimize $\{u_r,u_e,u_c\}$ with SGD\;
  }
\textbf{return}  learned ${f_\varphi}$ and ${f_M}$.}
\end{algorithm}
\vspace{-4mm}

Finally, we optimize $\mathcal{L}_M$ and $\mathcal{L}_y$ separately, of which the algorithm is shown in Alg. \ref{alg:AMTNet}. 
\seasons{Especially, we do not optimize the learnable variables $u$ (Line 7 of Alg. \ref{alg:AMTNet}) and network weights $f$ (Line 6 of Alg. \ref{alg:AMTNet}) simultaneously, because when $f$ changes, the competition of different metrics changes as well.}
Consequently, AMM is able to learn a generalized embedding via optimizing $\mathcal{L}_M$ based on multi-task collaborative learning paradigm, and obtain a comprehensive similarity measurement through optimizing $\mathcal{L}_y$.

\vspace{-4mm}
\section{Adaptive Metrics and Tasks Network}
\label{sec:amtnet}
On the basis of AMM, we design a novel training framework for FSL, named as \textit{Adaptive Metrics and Tasks Network} (AMTNet). The structure of the framework is illustrated in Fig.~\ref{fig:AMTNet}(a), {\color{black}which consists of the main few-shot branch (see the top in Fig.~\ref{fig:AMTNet}(a)) and the auxiliary self-supervision branch (see the bottom in Fig.~\ref{fig:AMTNet}(a)).
In training stage, a \textit{Global Adaptive Loss} (GAL) is proposed to coordinate the relationship between different losses generated from the above two branches. It also adopts the methodology of multi-task learning to optimize the model in an end-to-end manner, whose pseudo code is shown as in Algorithm \ref{alg:AMTNet}.} 
In inductive inference, for a task with novel data, the pre-trained embedding is directly utilized to extract the features of the support classes and query samples. 
Then the overall prediction ${\seasons{\hat{Y}}}$ for a query is predicted by the AMM based metric classifier via Eq.~\ref{equ:AMM_Y}.

\vspace{-4mm}
\subsection{Model Training via Optimization}
\label{sec:optimization}
As shown in Fig.~\ref{fig:AMTNet}, firstly each query sample $x^q$ is rotated under four angles $[0^{\circ}, 90^{\circ}, 180^{\circ}, 270^{\circ}]$, thus the query set $\mathcal{Q}=\{\left(x^q_i, y^q_i\right)\}_{i=1}^{n_q}$ is transformed into a rotated query set $\mathcal{\bar{Q}}=\{\left(\bar{x}^q_i, \bar{y}^q_i\right)\}_{i=1}^{{n_q}\times4}$. Then, the embedding ${f_\varphi}$ transfers the support set $\mathcal{S}$ and a rotated query sample $\bar{x}^q$ into the class prototype feature map $P^k\!=\!\frac{1}{|\mathcal{S}^k|}\!\sum_{x^s_i\in \mathcal{S}^k} {f_\varphi}(x^s_i)$ and a query feature map ${Q}={f_\varphi}(\bar{x}^q)$, respectively. {\color{black}Finally, AMTNet is optimized by minimizing the overall classification loss ${\mathcal{L}}$ contributing from the fused Metric $\mathcal{L}_M$ (defined in Eq. \ref{equ:AMM_LM}), the Global classifier ($\mathcal{L}_G$) and a Rotation classifier ($\mathcal{L}_R$), where the latter two are defined below.}

% As shown in Fig.~\ref{fig:AMTNet}, firstly each query sample $x^q$ is rotated under four angles [0$\degree$, 90$\degree$, 180$\degree$, 270$\degree$], thus the query set $\mathcal{Q}=\{\left(x^q_i, y^q_i\right)\}_{i=1}^{n_q}$ is transformed into a rotated query set $\mathcal{\bar{Q}}=\{\left(\bar{x}^q_i, \bar{y}^q_i\right)\}_{i=1}^{{n_q}\times4}$. Then, the embedding ${f_\varphi}$ transfers the support set $\mathcal{S}$ and a rotated query sample $\bar{x}^q$ into the class prototype feature map $P^k=\frac{1}{|\mathcal{S}^k|}\sum_{x^s_i\in \mathcal{S}^k} {f_\varphi}(x^s_i)$ and a query feature map ${Q}={f_\varphi}(\bar{x}^q)$, respectively. {\color{black}Finally, AMTNet is optimized by minimizing the overall classification loss ${\mathcal{L}}$ contributing from the fused Metric $\mathcal{L}_M$ (defined in Eq. \ref{equ:AMM_LM}), the Global classifier ($\mathcal{L}_G$) and a Rotation classifier ($\mathcal{L}_R$), where the latter two are defined below.}

{\color{black}The Metric classifier $f_M$ (see Fig.~\ref{fig:AMTNet}(a)) categorizes} the query images into $N$ support classes based on the proposed AMM similarity measurement. The Global classifier $f_G$ categorizes the query samples into all available classes of training set, and its loss is $\mathcal{L}_G = CE(\seasons{\hat{y}}_G, C^q)$, where $\seasons{\hat{y}}_G$ is the prediction results of the Global classifier, and ${C^q}$ is the true global class of $\bar{x}^q$ with total of $C$ classes. Similarly, the loss of Rotation classifier $f_R$ is computed as ${\mathcal{L}_R=CE(\seasons{\hat{y}}_R,B^q)}$, where $\seasons{\hat{y}}_R$ is the predictions, and ${B^q}$ is the true rotation class of $\bar{x}^q$ in {\color{black}four kinds of angle.}

\noindent\textbf{Global Adaptive Loss:} 
Then, the generic \textit{Global Adaptive Loss} (GAL) with three inputs $\{{\mathcal{L}_M},{\mathcal{L}_G},{\mathcal{L}_R}\}$ is defined as follows:
\vspace{-0.5mm}
\begin{equation}
\mathcal{L} = {w_M}{\mathcal{L}_M} + {w_G}{\mathcal{L}_G} + {w_R}{\mathcal{L}_R},
\label{equ:Loss_def}
\end{equation}
%\vspace{-1mm}
where \{${w_M,w_G,w_R}$\} are used to re-weight the losses of different tasks in optimizing.
According to \cite{hou2019cross}, it sets ${w_M=\frac{1}{2}}$, and recommends that the weight of auxiliary loss should be larger than metric loss (i.e. $\{w_G,w_R\}>w_M$). Further inspired by the multi-task learning defined in Eq.~\ref{equ:AMM_Lm}, we derive:
\vspace{-0.5mm}
\begin{equation}
\begin{aligned}
\mathcal{L} 
= \frac{1}{2}{\mathcal{L}_M} + \sum_{z=G,R}\left( w_z{\mathcal{L}_z}+{log{\frac{1}{w_z}}}\right),\
where \quad w_z={\frac{1}{{\theta_z^2}}+\lambda},
\end{aligned}
\label{equ:Loss}
\end{equation}
%\vspace{-1mm}
where, ${\mathcal{L}_M}$ is defined in Eq.~\ref{equ:AMM_LM}, learnable variables include \{${\theta_r}$, ${\theta_e}$, ${\theta_c}$\} and \{${\theta_G}$, ${\theta_R}$\}, and ${\lambda}$ is a hyper-parameter ({\color{black}empirical value is within [0.5, 2.0] to balance the effects of losses of few-shot and auxiliary tasks, as illustrated in Tab.~\ref{table:ablation_2}}). 
In Eq.~\ref{equ:Loss}, in order to approximate the condition of $\{w_G,w_R\}>w_M$, we propose to introduce a parameter ${\lambda}$ as the base bias for auxiliary loss. 
The results in Tab.~\ref{table:ablation_2} demonstrate that the proposed $\lambda$ plays an important role in final performance.

\vspace{-2mm}
\subsection{Discussion}
\noindent\textbf{The Benefits of GAL:}
In previous works \cite{gidaris2019boosting,su2020does,zhengyu2021pareto}, self-supervised rotation task is treated as an independent rotation prediction task, which only benefits embedding with rotation aware property.
{\color{black}The usage in our framework is obviously different as follows:} (i) Our approach exploits both rotation aware and rotation invariant properties, which results in a more semantic and robust embedding. 
Concretely, as illustrated in Fig.~\ref{fig:AMTNet}(a), Rotation classifier is rotation aware, and Global classifier and Metric classifier are rotation invariant. 
(ii) Different from these two stage embedding-based methods \cite{rizve2021exploring,zhengyu2021pareto}, our model is training in an end-to-end one stage manner (i.e. optimizes few-shot task and auxiliary tasks simultaneously).

\noindent\textbf{AMTNet with Knowledge Distillation:}
Our AMTNet is optimized via the proposed GAL based on multi-task paradigm, which is helpful to prevent overfitting and learn robust embedding.
With the usage of GAL, our AMTNet is able to achieve better few-shot classification performance with a larger backbone ${f_\varphi}$ such as WRN-28 instead of the ResNet-12.
As shown in Fig.~\ref{fig:AMTNet}(b), with knowledge distillation \cite{hinton2015distilling}, AMTNet+KD uses a strong teacher-AMTNet (WRN-28) model to train a student-AMTNet (ResNet-12).
The teacher-AMTNet is first trained via Algorithm \ref{alg:AMTNet}, then the student-AMTNet is optimized by the overall loss:
\vspace{-0.5mm}
\begin{equation}
\begin{aligned}
\mathcal{L} &= \mathcal{L}_{GAL} + \beta \mathcal{L}_{KD}, \quad where,\\
 \mathcal{L}_{GAL} &= \frac{1}{2}{\sum_{j=r,e,c}({{\frac{1}{{\theta_j^2}}}{\mathcal{L}_j}+{log{\theta_j^2}}})} \\ &\quad +  \sum_{z=G,R}\left({\left({\frac{1}{{\theta_z^2}}+\lambda}\right){\mathcal{L}_z}+{log{\frac{1}{{\left({\frac{1}{{\theta_z^2}}+\lambda}\right)}}}}}\right), \\
\mathcal{L}_{KD} &= \sum_{j=r,e,c}KL\left(\seasons{\hat{Y}}_j,\| \seasons{\hat{Y}}^{t}\|\right) + \sum_{z=G,R}KL\left(\seasons{\hat{Y}}_z,\seasons{\hat{Y}}^{t}_z\right),
\end{aligned}
\label{equ:Loss_kd}
\end{equation}
where $\beta$ is a hyper-parameter, $\seasons{\hat{Y}}^{t}$ and $\seasons{\hat{Y}}^{t}_z$ are the output predictions of teacher-AMTNet.

\renewcommand{\tabcolsep}{10pt}
\begin{table*}[t]
\caption{Comparison with SOTAs on 5-way classification on \emph{mini}ImageNet and \emph{tiered}ImageNet datasets. 
* indicates methods evaluated using \emph{multi-cropping}.
The best two results under different backbones are highlighted in \textbf{boldtype} and \textit{italics}.} 
\centering
\vspace{-0.2cm}
\begin{tabular}{ l | c | c | c c | c c}
\hline
\multicolumn{1}{l|}{\multirow{2}*{Model}}  & \multirow{2}*{Backbone} & \multirow{2}*{Venue} & \multicolumn{2}{c|}{\emph{mini}ImageNet}  &\multicolumn{2}{c}{\emph{tiered}ImageNet} \\   
\cline{4-7}
\multicolumn{1}{c|}{ } & & & 1-shot &5-shot &1-shot &5-shot \\
\hline
MatchingNet \cite{vinyals2016matching} &Conv4 &NeurIPS'2016 &43.44 $\pm$ 0.77 & 60.60 $\pm$ 0.71 & - & -\\
%MAML~\cite{finn2017model} &Conv4&ICML'2017 &48.70 $\pm$ 0.84 & 55.31 $\pm$ 0.73 & 51.67 $\pm$ 1.81 & 70.30 $\pm$ 1.75  \\
%MetaNet \cite{munkhdalai2017meta} &Conv4&ICML'2017 & 49.21 $\pm$ 0.96 & - & - &- \\
ProtoNet \cite{snell2017prototypical} &Conv4&NeurIPS'2017 &49.42 $\pm$ 0.78 & \textit{68.20 $\pm$ 0.66} &53.31 $\pm$ 0.89 &\textit{72.69 $\pm$ 0.74}\\
RelationNet \cite{sung2018learning} &Conv4&CVPR'2018 &\textit{50.44  $\pm$ 0.82} & 65.32  $\pm$ 0.70 & \textit{54.48  $\pm$ 0.93} & 71.32  $\pm$ 0.78\\
%MM-Net \cite{qi2018memory} &Conv4&CVPR'2018 &53.37 $\pm$ 0.48 &66.97 $\pm$ 0.35 & - & -\\
%\hdashline
\hline
\textbf{Our AMTNet} &Conv4&Ours & \textbf{54.91 $\pm$ 0.47} & \textbf{71.01 $\pm$ 0.37} & \textbf{57.33 $\pm$ 0.50} & \textbf{73.11 $\pm$ 0.36} \\
\hline
\hline
%adaNet \cite{munkhdalai2018rapid} &ResNet-12&ICML'2018 &56.88 $\pm$ 0.62 & 71.94 $\pm$ 0.57 & - &-\\
%TADAM~\cite{oreshkin2018tadam} &ResNet-12&NeurIPS'2018 &58.50 $\pm$ 0.30 & 76.70 $\pm$ 0.30 &- & -\\
%MTL~\cite{sun2019meta} &ResNet-12&CVPR'2019 &61.20 $\pm$ 1.80 & 75.50 $\pm$ 0.80 &- &- \\
%%MetaOpt~\cite{lee2019meta} &ResNet-12&CVPR'2019 &62.64 $\pm$ 0.62 &78.63 $\pm$ 0.46 & 65.99 $\pm$ 0.72 &81.56 $\pm$ 0.53\\
CAN~\cite{hou2019cross} &ResNet-12&NeurIPS'2019 &63.85 $\pm$ 0.48 & 79.44 $\pm$ 0.34 &69.89 $\pm$ 0.51 &84.23 $\pm$ 0.37 \\
%%P-Transfer~\cite{zhiqiang2021partial} &ResNet-12&AAAI'2021 &64.21 $\pm$ 0.77 & 80.38 $\pm$ 0.59 &- &- \\
%MetaOpt+ArL~\cite{hongguang2021rethink}&ResNet-12&CVPR'2021 &65.21 $\pm$ 0.58 &80.41 $\pm$ 0.49 &- &-\\
DeepEMD~\cite{zhang2020deepemd} &ResNet-12&CVPR'2020 &65.91 $\pm$ 0.82 & 82.41 $\pm$ 0.56 &71.16 $\pm$ 0.87 &{86.03 $\pm$ 0.58} \\
IENet~\cite{rizve2021exploring} &ResNet-12&CVPR'2021 &{66.82 $\pm$ 0.80} & \textit{84.35 $\pm$ 0.51} &{71.87 $\pm$ 0.89} &{86.82 $\pm$ 0.58} \\
infoPatch~\cite{liu2021learning} &ResNet-12&AAAI'2021 &67.67 $\pm$ 0.45 & 82.44 $\pm$ 0.31 &71.51 $\pm$ 0.52 &85.44 $\pm$ 0.35 \\
RFS~\cite{tian2020rethinking} &ResNet-12&ECCV'2020 &67.73 $\pm$ 0.63 & 83.35 $\pm$ 0.41 &72.55 $\pm$ 0.69 &86.72 $\pm$ 0.49 \\
DANet~\cite{xu2021learning} &ResNet-12&CVPR'2021 &67.76 $\pm$ 0.46 & 82.71 $\pm$ 0.31 &71.89 $\pm$ 0.52 &85.96 $\pm$ 0.35 \\
COSOC* \cite{xu2021Rectifying} &ResNet-12&NeurIPS'2021 &\textit{69.28 $\pm$ 0.49} & \textbf{85.16 $\pm$ 0.42} &\textbf{73.57 $\pm$ 0.43} &\textbf{87.57 $\pm$ 0.10} \\
%\hdashline
\hline
\textbf{Our AMTNet} &ResNet-12&Ours &{69.17 $\pm$ 0.46} &{83.88 $\pm$ 0.31} & {72.63 $\pm$ 0.49} & 86.54 $\pm$ 0.36 \\
\textbf{Our AMTNet+KD} &ResNet-12&Ours &\textbf{69.48 $\pm$ 0.47} &{84.22 $\pm$ 0.31} & \textit{73.02 $\pm$ 0.50} & \textit{86.98 $\pm$ 0.36} \\
\hline
\hline
wDAE-GNN~\cite{gidaris2019generating} &WRN-28&CVPR'2019 &61.07 $\pm$ 0.15 &76.75 $\pm$ 0.11 &68.18 $\pm$ 0.16 & 83.09 $\pm$ 0.12 \\
LEO~\cite{rusu2019meta} &WRN-28&ICLR'2019 &61.76 $\pm$ 0.08 &77.59 $\pm$ 0.12 & 66.33 $\pm$ 0.05 & 81.44 $\pm$ 0.09 \\
wDAE~\cite{gidaris2019generating} &WRN-28&CVPR'2019 &62.96 $\pm$ 0.15 & 78.85 $\pm$ 0.10 & 68.18 $\pm$ 0.16 & 83.09 $\pm$ 0.12 \\
PSST~\cite{zhengyu2021pareto} &WRN-28&CVPR'2021 &64.16 $\pm$ 0.44 & 80.64 $\pm$ 0.32 &- &- \\
%SimpleShot~\cite{yan2019simpleshot} &DenseNet-121&arXiv'2019 &64.29 $\pm$ 0.20 & 81.50 $\pm$ 0.14 &71.32 $\pm$ 0.22 &86.66 $\pm$ 0.15 \\
FEAT~\cite{ye2020few} &WRN-28&CVPR'2020 &65.10 $\pm$ 0.20 & 81.11 $\pm$ 0.14 &70.41 $\pm$ 0.23 &84.38 $\pm$ 0.16 \\
%CA~\cite{afrasiyabi2019associative} &WRN-28&arXiv'2019 &65.92 $\pm$ 0.60 & 82.85 $\pm$ 0.55 &\textbf{74.40 $\pm$ 0.68} &86.61 $\pm$ 0.59 \\
CAN~\cite{hou2019cross} &WRN-28&NeurIPS'2019 &66.12 $\pm$ 0.47 & 80.43 $\pm$ 0.33 &71.04 $\pm$ 0.53 &84.92 $\pm$ 0.37 \\
DANet~\cite{xu2021learning} &WRN-28&CVPR'2021 &\textit{67.84 $\pm$ 0.46} & \textit{82.74 $\pm$ 0.31} &\textit{72.18 $\pm$ 0.52} &\textit{86.26 $\pm$ 0.35} \\
%\hdashline
\hline
\textbf{Our AMTNet} &WRN-28& Ours& \textbf{70.05 $\pm$ 0.46} &\textbf{84.55 $\pm$ 0.29} &\textbf{73.86 $\pm$ 0.50} &\textbf{87.62 $\pm$ 0.33} \\
\hline
\end{tabular}
\vspace{-0.2cm}
\label{table:SOTA}
\end{table*}

\renewcommand{\tabcolsep}{3pt}
\begin{table}[t]
%\caption{Comparison with SOTAs on 5-way classification on CIFAR-FS dataset.} 
\caption{Comparison on 5-way task on CIFAR-FS dataset.} 
\centering
\vspace{-0.2cm}
\begin{tabular}{ l | c | c c}
\hline
\multicolumn{1}{l|}{\multirow{2}*{Model}}  & \multirow{2}*{Backbone} & \multicolumn{2}{c}{CIFAR-FS} \\   
\cline{3-4}
\multicolumn{1}{c|}{ } & & 1-shot &5-shot \\
\hline
RFS~\cite{tian2020rethinking} &ResNet-12 &71.50 $\pm$ 0.80 &86.00 $\pm$ 0.50 \\
MetaOpt~\cite{lee2019meta} &ResNet-12 &72.60 $\pm$ 0.70 &84.30 $\pm$ 0.50 \\
MABAS~\cite{jaekyeom2020model} &ResNet-12 &73.51 $\pm$ 0.92 &85.49 $\pm$ 0.68 \\
DSN-MR~\cite{simon2020adaptive} &ResNet-12 &75.60 $\pm$ 0.90 &86.20 $\pm$ 0.60 \\
IENet~\cite{rizve2021exploring} &ResNet-12 &76.83 $\pm$ 0.82 &89.26 $\pm$ 0.58 \\
%\hdashline
\hline
\textbf{Our AMTNet} &ResNet-12 &78.94 $\pm$ 0.48 &89.28 $\pm$ 0.33 \\
\textbf{Our AMTNet+KD} &ResNet-12 &\textbf{79.52 $\pm$ 0.48} &\textbf{89.60 $\pm$ 0.33} \\
\hline
\hline
\textbf{Boosting}~\cite{gidaris2019boosting} &WRN-28 &73.60 $\pm$ 0.30 &86.00 $\pm$ 0.20 \\
\textbf{Fine-tuning}~\cite{dhillon2019baseline} &WRN-28&76.58 $\pm$ 0.68 &85.79 $\pm$ 0.50 \\
%\hdashline
\hline
\textbf{Our AMTNet} &WRN-28&\textbf{80.38 $\pm$ 0.48} &\textbf{89.89 $\pm$ 0.32} \\
\hline
\end{tabular}
\vspace{-0.4cm}
\label{table:SOTA_cifar}
\end{table}

\renewcommand{\tabcolsep}{3.5pt}
\begin{table}[ht]
\caption{The results on 5-way classification with different metric methods under ResNet-12 backbone. The Metric and Global loss weights are set to 0.5 and 1.0 respectively, and the Rotation classifier is not applied.}
\centering
\vspace{-0.2cm}
\begin{tabular}{l | c | c |  c | c  c}
\hline
\multirow{2}*{Metric}  &\multirow{2}*{$\seasons{\hat{Y}}$}  &\multirow{2}*{$\mathcal{L}_M$}  & \multirow{2}*{Param} & \multicolumn{2}{c}{\emph{mini}ImageNet} \\   
\cline{5-6}
 & & & & 1-shot &5-shot \\
\hline
Relation &$\seasons{\hat{Y}}_r$ &$\mathcal{L}_r$  &8.50M &61.84 $\pm$ 0.48 &77.48 $\pm$ 0.35  \\
Euclidean &$\seasons{\hat{Y}}_e$ &$\mathcal{L}_e$  &7.75M &62.90 $\pm$ 0.48 &78.03 $\pm$ 0.35  \\
Cosine &$\seasons{\hat{Y}}_c$ &$\mathcal{L}_c$ &7.75M &64.45 $\pm$ 0.47 &79.20 $\pm$ 0.34  \\
%\hdashline
\hline
Coupled &Eq.~\ref{equ:pf_nmm} &$\mathcal{L}_y$  &8.50M &66.04 $\pm$ 0.47 &80.03 $\pm$ 0.34  \\
NMM &Eq.~\ref{equ:pf_nmm} &Eq.~\ref{equ:lf_nmm} &8.50M &66.48 $\pm$ 0.47 &80.26 $\pm$ 0.34  \\
AMM-V1 &Eq.~\ref{equ:pf_nmm} &Eq.~\ref{equ:AMM_Lm} &8.50M &67.50 $\pm$ 0.47 &81.18 $\pm$ 0.33  \\
AMM-V2 &Eq.~\ref{equ:AMM_Y} &Eq.~\ref{equ:AMM_Lm} &8.50M &{68.02 $\pm$ 0.46} &{82.64 $\pm$ 0.32}  \\
AMM &Eq.~\ref{equ:AMM_Y} &Eq.~\ref{equ:AMM_LM} &8.50M &\textbf{68.47 $\pm$ 0.46} &\textbf{83.23 $\pm$ 0.31}  \\
\hline
\end{tabular}
\vspace{-0.4cm}
\label{table:ablation_1}
\end{table}

\vspace{-4mm}
\section{Experiment}
\subsection{Experimental Setup}
\noindent\textbf{Datasets:}
\textbf{\emph{mini}ImageNet} dataset is a subset of ImageNet \cite{krizhevsky2012imagenet}, which consists of 100 classes with image size of 84${\times}$84 pixels. 
We split the 100 classes following the setting in \cite{xu2021learning}, i.e. 64, 16 and 20 classes for training, validation and testing respectively. 
\textbf{\emph{tiered}ImageNet} dataset \cite{ren2018meta} is also a subcollection of ImageNet. It contains 608 classes with image size of 84${\times}$84 pixels, which are separated into 351 classes for training, 97 for validation and 160 for testing.
\textbf{CIFAR-FS} dataset is constructed by randomly splitting the 100 classes of the CIFAR-100 dataset into 64, 16, and 20 train, validation, and test splits.

\noindent\textbf{Evaluation:}
We conduct experiments on $5$-way $1$-shot and $5$-shot inductive settings. We report the \textit{average accuracy} and $95\%$ \textit{confidence interval} over $2000$ episodes sampled from the test set. 
%\ljx{In addition to the inductive FSL setting,  transductive setting \cite{yanbin2018learn,pau2020embed,jie2021rein,baoquan2021proto} utilizes extra unlabeled data in inference, which is \seasons{similar to} few-shot semi-supervised learning. 
%\seasons{Here}, we do not adopt transductive setting.}
\ljx{Here, we do not adopt transductive setting \cite{yanbin2018learn,pau2020embed,jie2021rein,baoquan2021proto}.}

\noindent\textbf{Implementation details:}
Following \cite{xu2021learning}, horizontal flipping, random cropping, random erasing and color jittering are employed for data augmentation in training.
According to the ablation study results, the hyperparameter $\lambda$ in Eq.~\ref{equ:Loss} is set to $0.5$ and $1.5$ under ResNet-12 and WRN-28 respectively, and $\beta$ in Eq.~\ref{equ:Loss_kd} is set to $0.75$ for AMTNet+KD. 
In line with the setting of \cite{hou2019cross}, SGD with $5e$-$4$ weight decay is used as the optimizer.
The detailed info of learning-rate and training-epochs are referred to our public source code.

\vspace{-4mm}
\subsection{Comparison with State-of-the-arts}
Tab.~\ref{table:SOTA} and Tab.~\ref{table:SOTA_cifar} compare our method with existing few-shot methods on \emph{mini}ImageNet, \emph{tiered}ImageNet and CIFAR-FS, which indicates that the proposed AMTNet outperforms the existing SOTAs with a large margin under WRN-28 backbone as well as is very competitive under Conv4 and ResNet-12. 

On \emph{mini}ImageNet, our AMTNet performs better than the best parameter-generating method wDAE~\cite{gidaris2019generating} with an improvement up to $7.09\%$.
Comparing to Pareto Optimal multi-task collaborative learning based PSST~\cite{zhengyu2021pareto} approach, our GAL based AMTNet achieves $5.89\%$ higher performance.
Many existing metric-based methods~\cite{vinyals2016matching,snell2017prototypical,sung2018learning,zhang2020deepemd} focus on designing different distance metrics for few-shot classification, and the strongest competitor is DeepEMD~\cite{zhang2020deepemd} with Earth Mover’s Distance. Our AMTNet is $3.57\%$ higher than DeepEMD, which demonstrates the superiority of the proposed adaptive metrics module. 
Some metric-based methods~\cite{hou2019cross,xu2021learning} apply cross attention strategy to get more discriminative features before metric classification. Even without any feature attention, our method still outperforms the DANet~\cite{xu2021learning} with an improvement up to $1.72\%$.

Besides, COSOC \cite{xu2021Rectifying} method adopts a complicated multi-stage framework, including pre-training backbone by contrastive learning, data clustering, generating cropped data, training backbone by FSL algorithm and inference with image-cropping. Furthermore, COSOC also takes an extra cost at inference step, because it needs to crop the source image several times for similarity calculation.
Comparing to COSOC, our AMTNet is an end-to-end framework, which achieves new state-of-the-art results with WRN-28.
%With the usage of knowledge distillation, the AMTNet+KD obtains higher accuracy on 1-shot task under ResNet-12.

\renewcommand{\tabcolsep}{4.5pt}
\begin{table}[t]
\caption{The results on 5-way classification about the influence of the hyper-parameter $\alpha$ as introduced in Eq.~\ref{equ:AMM_LM}.
The setting is consistent with Tab.~\ref{table:ablation_1}.}
\centering
\vspace{-0.1cm}
\begin{tabular}{l | c | c | c | c  c}
\hline
\multirow{2}*{Metric}  &\multirow{2}*{$\seasons{\hat{Y}}$}  &\multirow{2}*{$\mathcal{L}_M$}  & \multirow{2}*{$\alpha$} & \multicolumn{2}{c}{\emph{mini}ImageNet} \\   
\cline{5-6}
 & & & & 1-shot &5-shot \\
\hline
AMM-V2 &Eq.~\ref{equ:AMM_Y} &Eq.~\ref{equ:AMM_Lm} &- &{68.02 $\pm$ 0.46} &{82.64 $\pm$ 0.32}  \\
%\hdashline
\hline
AMM &Eq.~\ref{equ:AMM_Y} &Eq.~\ref{equ:AMM_LM} &1.0 &{68.06 $\pm$ 0.47} &{83.11 $\pm$ 0.32}  \\
AMM &Eq.~\ref{equ:AMM_Y} &Eq.~\ref{equ:AMM_LM} &0.5 &{68.18 $\pm$ 0.46} &\textbf{83.23 $\pm$ 0.31}  \\
AMM &Eq.~\ref{equ:AMM_Y} &Eq.~\ref{equ:AMM_LM} &0.1 &\textbf{68.47 $\pm$ 0.46} &{82.89 $\pm$ 0.32}  \\
\hline
\end{tabular}
\vspace{-0.4cm}
\label{table:ablation_kl}
\end{table}

\renewcommand{\tabcolsep}{8.5pt}
\begin{table*}[ht]
\caption{The results on 5-way classification on \emph{mini}ImageNet about the influence of Global Adaptive Loss (GAL) employed in AMTNet under ResNet-12 and WRN-28 backbones. There are two experimental groups: \textbf{Group1}, showing the influence of Global and Rotation tasks. \textbf{Group2}, searching the experimental optimal hyper-parameter ${\lambda}$ of GAL as introduced in Eq.~\ref{equ:Loss}.
}
\centering
\begin{tabular}{l | c | c c c | c  c | c  c}
\hline
\multirow{2}*{Exp. group}&\multirow{2}*{${\lambda}$} & \multicolumn{3}{c|}{Loss weights} & \multicolumn{2}{c|}{ResNet-12} & \multicolumn{2}{c}{WRN-28} \\   
\cline{3-9}
&&Metric &Global &Rotation & 1-shot &5-shot & 1-shot &5-shot \\
\hline
\multirow{4}*{\textbf{Group1}}
&-&0.5 &- &- &62.43 $\pm$ 0.50 &80.61 $\pm$ 0.33 &61.15 $\pm$ 0.50 &76.90 $\pm$ 0.38 \\
&-&0.5 &- &1.0 &65.33 $\pm$ 0.50 &80.43 $\pm$ 0.35 &63.84 $\pm$ 0.50 &78.50 $\pm$ 0.38 \\
&-&0.5 &1.0 &- &68.47 $\pm$ 0.46 &83.23 $\pm$ 0.31 &67.07 $\pm$ 0.48 &82.59 $\pm$ 0.31 \\
&-&0.5 &1.0 &1.0 &\textbf{68.80 $\pm$ 0.47} &\textbf{83.77 $\pm$ 0.30} &\textbf{69.07 $\pm$ 0.48} &\textbf{84.36 $\pm$ 0.29} \\
\hline
\multirow{8}*{\textbf{Group2}}
&0.0&0.5 &${w_G}$ &${w_R}$ &67.91 $\pm$ 0.48 &83.15 $\pm$ 0.31 &67.32 $\pm$ 0.50 &82.59 $\pm$ 0.32 \\
&0.5&0.5 &${w_G}$ &${w_R}$ &\textbf{69.17 $\pm$ 0.46} &\textbf{83.88 $\pm$ 0.31} &68.61 $\pm$ 0.48 &84.01 $\pm$ 0.30 \\
&1.0&0.5 &${w_G}$ &${w_R}$ &68.94 $\pm$ 0.46 &83.76 $\pm$ 0.31 &69.32 $\pm$ 0.47 &84.14 $\pm$ 0.31 \\
&1.5&0.5 &${w_G}$ &${w_R}$ &68.53 $\pm$ 0.46 &82.97 $\pm$ 0.31 &\textbf{70.05 $\pm$ 0.46} &\textbf{84.55 $\pm$ 0.29}  \\
&2.0&0.5 &${w_G}$ &${w_R}$ &67.95 $\pm$ 0.46 &82.76 $\pm$ 0.31 &69.85 $\pm$ 0.46 &84.52 $\pm$ 0.29 \\
&3.0&0.5 &${w_G}$ &${w_R}$ &66.94 $\pm$ 0.46 &82.11 $\pm$ 0.32 &69.93 $\pm$ 0.46 &84.51 $\pm$ 0.29 \\
&4.0&0.5 &${w_G}$ &${w_R}$ &66.31 $\pm$ 0.46 &81.39 $\pm$ 0.32 &69.25 $\pm$ 0.46 &84.33 $\pm$ 0.30 \\
&6.0&0.5 &${w_G}$ &${w_R}$ &61.45 $\pm$ 0.47 &81.02 $\pm$ 0.33 &69.62 $\pm$ 0.46 &83.63 $\pm$ 0.30 \\
\hline
\end{tabular}
\vspace{-0.0cm}
\label{table:ablation_2}
\end{table*}

\renewcommand{\tabcolsep}{3.5pt}
\begin{table}[t]
\caption{The results of AMTNet+KD on 5-way classification.
The hyper-parameter $\beta$ is introduced in Eq.~\ref{equ:Loss_kd}.
}
\centering
\vspace{-0.2cm}
\begin{tabular}{c | c | c | c | c  c}
\hline
\multirow{2}*{AMTNet}&\multirow{2}*{Backbone}&\multirow{2}*{${\beta}$}&\multirow{2}*{${\lambda}$} & \multicolumn{2}{c}{\emph{mini}ImageNet}\\
\cline{5-6}
&&& & 1-shot &5-shot \\
\hline
Teacher
&ResNet-12&-&0.5 &{69.17 $\pm$ 0.46} &{83.88 $\pm$ 0.31} \\
%\hdashline
\hline
\multirow{5}*{{Student}}
&\multirow{5}*{{ResNet-12}}&1.0&0.5 &{69.06 $\pm$ 0.47} &{83.61 $\pm$ 0.31} \\
&&0.75&0.5&68.96 $\pm$ 0.47 &83.86 $\pm$ 0.31  \\
&&0.5&0.5&\textbf{69.28 $\pm$ 0.47} &\textbf{83.99 $\pm$ 0.31} \\
&&0.25&0.5&{69.22 $\pm$ 0.47} &{83.71 $\pm$ 0.31} \\
&&0.1&0.5&{69.16 $\pm$ 0.47} &{83.57 $\pm$ 0.31} \\
\hline
\hline
Teacher
&WRN-28&-&1.5 &{70.05 $\pm$ 0.46} &{84.55 $\pm$ 0.29} \\
%\hdashline
\hline
\multirow{5}*{{Student}}
&\multirow{5}*{{ResNet-12}}&1.0&0.5 &{68.59 $\pm$ 0.47} &{83.72 $\pm$ 0.31} \\
&&0.75&0.5&\textbf{69.48 $\pm$ 0.47} &\textbf{84.22 $\pm$ 0.31}  \\
&&0.5&0.5&69.26 $\pm$ 0.46 &83.84 $\pm$ 0.31 \\
&&0.25&0.5&{69.13 $\pm$ 0.47} &{83.87 $\pm$ 0.31} \\
&&0.1&0.5&{69.27 $\pm$ 0.46} &{83.84 $\pm$ 0.31} \\
\hline
\end{tabular}
\vspace{-0.4cm}
\label{table:ablation_kd}
\end{table}

\subsection{Model Analysis}
\noindent\textbf{Effectiveness of Adaptive Metrics Module:}
In Tab.~\ref{table:ablation_1}, different metric methods are compared without rotation task \seasons{using Res-12 backbone}.
The proposed AMM is 4.02\% and 4.03\% higher than the best individual Cosine metric on 1-shot and 5-shot tasks, respectively. Comparing to any individual metric method, NMM achieves impressive accuracy gains which indicates increasing the metric diversity boosts the performance.
And AMM gains a further accuracy improvement around 2\% upon NMM via alleviating the metrics competition problem. 
\seasons{Due to length constraints, the results of backbone WRN-28 is shown in the APPENDIX, which indicates the similar performance compared with ResNet-12.}
%\seasons{Due to length constraints, the results of backbone WRN-28 is shown in the appendix, which indicates the similar performance compared with ResNet-12.}

Specifically, the results in Tab.~\ref{table:ablation_1} demonstrate the effectiveness of our approaches: (i) Comparing NMM to Coupled: decoupling metrics fusion into predictions fusion and losses fusion is helpful; (ii) Comparing AMM-V1 to NMM: multi-metrics loss fusion is able to learn a more robust embedding; (iii) Comparing AMM-V2 to AMM-V1: the adaptive layer controls the contributions of different metrics to alleviate metrics competition; (iv) Comparing AMM to AMM-V2: the KL regularization $\mathcal{L}_{KL}$ increases consistencies between different metric distributions to alleviate the metric criterion discordance problem.

Tab.~\ref{table:ablation_kl} shows the influence of the hyper-parameter $\alpha$ as introduced in Eq.~\ref{equ:AMM_LM}. 
The KL regularization $\mathcal{L}_{KL}$ considers AMM $\seasons{\hat{Y}}$ as the teacher-metric and individual metric $\seasons{\hat{Y}}_j$ as the student-metric.
The optimal values of $\alpha$ are 0.1 on 1-shot task and 0.5 on 5-shot task respectively, which indicates that AMM obtains a more stable teacher-metric on 5-shot task than on 1-shot task.

%\iffalse
\noindent\textbf{Influence of Global Adaptive Loss:}
As illustrated in Tab.~\ref{table:ablation_2}, our AMTNet obtains its best results as setting ${\lambda}$ to ${0.5}$ and ${1.5}$ under ResNet-12 and WRN-28 backbones respectively.
Based on the proposed AMM and GAL modules, AMTNet achieves large accuracy improvements (maximum up-to 8.9\%) on 1-shot and 5-shot tasks comparing to the model without GAL (first row in Tab.~\ref{table:ablation_2}). 
As shown in \textbf{Group2}, the recommended setting is ${\lambda \in [0.5, 2.0]}$, and our method under WRN-28 gets competitive performance on a large range of ${\lambda \in [1.0, 4.0]}$.
In \textbf{Group1}, the results indicate that the auxiliary tasks (i.e. Global classification and Rotation classification) are useful for training a more robust embedding leading to an accuracy improvement.

\noindent\textbf{Effectiveness of Knowledge Distillation:}
As illustrated in Tab.~\ref{table:ablation_kd}, we obtain a further accuracy improvements with the usage of knowledge distillation. The Student-AMTNet (ResNet-12) achieves its best results as setting ${\beta}$ to ${0.75}$ with Teacher-AMTNet (WRN-28).
Thus, our AMTNet+KD approach encourages adopting a large backbone (WRN-28) to achieve better accuracy performance, then using knowledge distillation technology to improve the accuracy on a smaller (ResNet-12) backbone.

\vspace{-2mm}
\section{Conclusion}
In this paper, we investigate the contributions of different distance metrics, and propose an effective few-shot classification framework, named AMTNet, which consists of two novel structures: \textit{Adaptive Metrics Module} (AMM) and the \textit{Global Adaptive Loss} (GAL). 
Specifically, AMM integrates both flexible and fixed distance metrics to achieve mutual complementarity in embedding learning and metric-decision making. 
To deal with the metrics competition problem, the proposed AMM decouples metrics fusion into predictions fusion and losses fusion, and further utilizes KL regularization to increase consistencies between different metric distributions. 
Moreover, GAL further riches the embedding by providing more supervised information from multiple tasks which gains a significant performance improvement. 
Extensive experiments show that our method is effective for few-shot classification, and achieves new state-of-the-art results on \emph{mini}ImageNet and \emph{tiered}ImageNet benchmark datasets.

\vspace{-2mm}
\section*{Acknowledgement}
This work was supported by the National Key Research and Development Program of China (2021ZD0111000), National Natural Science Foundation of China No. 62176092, Shanghai Science and Technology Commission No.21511100700, Natural Science Foundation of Shanghai (20ZR1417700).

%\newpage
\bibliographystyle{ACM-Reference-Format}
\balance
\bibliography{sample-base}

%\iffalse
%\newpage
\appendix

\section{appendix}
This section provides more details of our proposed method and experimental results, which are omitted in the main paper due to space limitation.

\subsection{Structure of the Applied Relation Module}
\begin{figure}[t]
\centering
\includegraphics[width=0.65\linewidth]{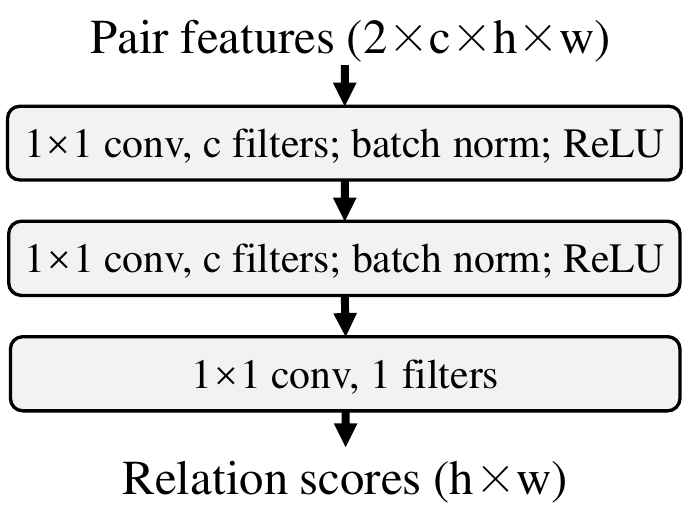}
\vspace{1mm}
\caption{The variant Relation module applied in AMM.}
\label{fig:RM}
\vspace{-3mm}
\end{figure}

The detail structure of the variant Relation Module adopted in AMTNet is shown in Fig.~\ref{fig:RM}. 

\subsection{Derivation of Losses Fusion in AMM}
\label{sec:lf_derive}
The detailed formula derivation of losses fusion as in Eq.\ref{equ:AMM_Lm} for AMM, is described as follow.
To realize task-dependent uncertainty in multi-task loss function \cite{alex2018multi}, the classification likelihood originally calculated by Eq.\ref{equ:pred} is now modified into a \textit{scaled} version:
\begin{equation}
\seasons{\hat{y}}_j(y=k|{Q};{\theta}_j) =\frac{\exp{\left(-{\frac{1}{{\theta_j^2}}} \cdot d_j\left({Q}, {P}^k\right)\right)}}
{\sum_{i=1}^{N} \exp{\left(-{\frac{1}{{\theta_j^2}}} \cdot d_j\left({Q}, {P}^i\right)\right)}},
\label{equ:pred_ap}
\end{equation}
where ${\theta}_j$ is a positive scalar of which the parameter's magnitude determines how `uniform' (flat) the discrete classification distribution is. This relates to its uncertainty, as measured in entropy.
It can be interpreted as a Boltzmann distribution where the input is scaled by $\theta_j^2$ (often referred as \textit{temperature}). 
Then the classification cross-entropy loss $\mathcal{L}_j(\seasons{\hat{y}}_j;{\theta}_j)$ for this output can be written as:
\begin{equation}
\begin{aligned}
&= -\log \seasons{\hat{y}}_j(y={y}^q_i|{Q}_{i};{\theta}_j) \\
&= -\log \frac{\exp{\left(-{\frac{1}{{\theta_j^2}}} \cdot d_j\left({Q}, {P}^k\right)\right)}}
{\sum_{i=1}^{N} \exp{\left(-{\frac{1}{{\theta_j^2}}} \cdot d_j\left({Q}, {P}^i\right)\right)}} \\
&= {\frac{1}{{\theta_j^2}}} \cdot d_j\left({Q}, {P}^k\right) + \log{\sum_{i=1}^{N} \exp{\left(-{\frac{1}{{\theta_j^2}}} \cdot d_j\left({Q}, {P}^i\right)\right)}}\\
&= -{\frac{1}{{\theta_j^2}}} \log \frac{\exp{\left(-d_j\left({Q}, {P}^k\right)\right)}}
{\sum_{i=1}^{N} \exp{\left(-d_j\left({Q}, {P}^i\right)\right)}}\\
& \qquad +\left(\log {\theta_j^2} \cdot \frac{\sum_{i=1}^{N} \exp{\left(-{\frac{1}{{\theta_j^2}}} \cdot d_j\left({Q}, {P}^i\right)\right)}} {\sum_{i=1}^{N} \exp{\left(-d_j\left({Q}, {P}^i\right)\right)}}\right)\\
&\approx {{\frac{1}{{\theta_j^2}}}{\mathcal{L}_j}+{log{\theta_j^2}}}.
\end{aligned}
\label{equ:Lj_ap}
\end{equation}
Therefore, based on Eq.\ref{equ:Lj_ap}, we obtain the multi-task loss for metric losses fusion as expressed in Eq.\ref{equ:AMM_Lm}.

\subsection{AMTNet Training with Patch-wise Strategy}
\label{sec:optimization_ap}
Supplement to Sec.\ref{sec:optimization}, the detailed objective functions for Metric, Global and Rotation classifiers are illustrated as follow.

To produce precise embeddings, each spatial position of the query features are constrained to be independently classified which is called as the patch-wise classification strategy as referred in \cite{hou2019cross}. For \textbf{Metric classifier}, each local feature $Q_n$ in the $n^{th}$ position of $Q$, is classified into $N$ support classes. Formally, the probability of predicting $Q_n$ as $k^{th}$ class is:
\begin{equation}
\seasons{\hat{y}}_j(y=k|Q_n)=\frac{\exp{\left(-d_j\left(Q_n, \textit{GAP}({P}^k)\right)\right)}}
{\sum_{i=1}^{N} \exp{\left(-d_j\left(Q_n, \textit{GAP}({P}^i)\right)\right)}},
\label{equ:p_pred}
\end{equation}
where \textit{GAP} represents global average pooling to obtain class feature. According to the true N-way few-shot class label $\bar{y}^q$, each individual metric classification loss is then defined as:
\begin{equation}
\mathcal{L}_j = -\sum_{i=1}^{n_q} \sum_{n=1}^{h \times w}\log \seasons{\hat{y}}_j(y=\bar{y}^q_i|(Q_{n})_{i}).
\label{equ:p_Lj}
\end{equation}

The \textbf{Global classifier} categorizes the query into all available classes of training set, and its loss is:
\begin{equation}
\begin{aligned}
\mathcal{L}_G&=PCE(Q,C^q) \\
&=-\sum_{i=1}^{n_q} \sum_{n=1}^{h \times w} {C^q_i} \log \left(\textit{softmax}(W(Q_n)_i)\right).
\end{aligned}
\label{equ:p_LG}
\end{equation}
where, ${PCE}$ is defined as the patch-wise cross-entropy function, and $W$ is the ${Linear}$ layer. Similarly, the loss of \textbf{Rotation classifier} is computed as ${\mathcal{L}_R=PCE(Q,B^q)}$.

\renewcommand{\tabcolsep}{5pt}
\begin{table}[bph]
\renewcommand\arraystretch{1.3}
\vspace{-0.2cm}
\caption{The results on 5-way 1-shot classification with different metric methods based on AMTNet framework with WRN-28 backbone.}\label{ablation_wrn}
\vspace{-0.3cm}
\centering
\begin{tabular}{ c | c | c | c | c | c}
\hline
Metric & AMM & NMM & Relation & Euclidean & Cosine   \\
\hline
Param & 36.25M & 36.25M & 36.25M & 35M & 35M  \\
\hline
1-shot & 70.05\% & 68.46\% & 65.12\% & 66.45\% & 67.67\%  \\
\hline
\end{tabular}
\end{table}
\vspace{-0.3cm}

%\fi

\end{document}